\documentclass[10pt,twocolumn,letterpaper,tikz]{article}

\usepackage{cvpr}
\usepackage{times}
\usepackage{epsfig}
\usepackage{graphicx}
\usepackage{amsmath}
\usepackage{amssymb}
\usepackage{tabularx}
\usepackage{array}
\usepackage{ctable}
\usepackage{tabulary}
\usepackage{multirow}
\usepackage[T1]{fontenc}
\usepackage{pifont}
\usepackage{subcaption}
\usepackage{enumitem}

\newcommand{\cmark}{\ding{51}}%
\newcommand{\xmark}{\ding{55}}%

\usepackage[pagebackref=true,breaklinks=true,colorlinks,bookmarks=false]{hyperref}

\cvprfinalcopy 

\def\cvprPaperID{****} 
\def\httilde{\mbox{\tt\raisebox{-.5ex}{\symbol{126}}}}

\ifcvprfinal\fi
\begin{document}

\title{OrigamiNet: Weakly-Supervised, Segmentation-Free, One-Step, Full Page Text Recognition by learning to unfold}

\author{Mohamed Yousef\\
Intuition Machines, Inc.\\
{\tt\small myb@imachines.com}
\and
Tom E. Bishop\\
Intuition Machines, Inc.\\
{\tt\small tom@imachines.com}
}

\maketitle

\begin{abstract}
   Text recognition is a major computer vision task with a big set of associated challenges. One of those traditional challenges is the coupled nature of text recognition and segmentation. This problem has been progressively solved over the past decades, going from segmentation based recognition to segmentation free approaches, which proved more accurate and much cheaper to annotate data for. We take a step from segmentation-free single line recognition towards segmentation-free multi-line / full page recognition. We propose a novel and simple neural network module, termed \textbf{OrigamiNet}, that can augment any CTC-trained, fully convolutional single line text recognizer, to convert it into a multi-line version by providing the model with enough spatial capacity to be able to properly collapse a 2D input signal into 1D without losing information. Such modified networks can be trained using exactly their same simple original procedure, and using only \textbf{unsegmented} image and text pairs. We carry out a set of interpretability experiments that show that our trained models learn an accurate implicit line segmentation. We achieve state-of-the-art character error rate on both IAM \& ICDAR 2017 HTR benchmarks for handwriting recognition, surpassing all other methods in the literature. On IAM we even surpass single line methods that use accurate localization information during training. Our code is available online at \small\url{https://github.com/IntuitionMachines/OrigamiNet}.
\end{abstract}

\section{Introduction}

\newcolumntype{C}[1]{>{\centering}m{#1}}

The ubiquity of text has made the automation of the processing of its various visual forms, an ever-increasing necessity. Over the years, one of the main driving themes for error rate reduction in text recognition systems has been reducing explicit segmentation proposals in favor of increasing full sequence recognition. In full sequence models, the recognition system learns to both simultaneously segment / align and recognize / classify an image representing a sequence of observations (i.e. characters). 
This trend progressed from the first systems that tried to segment each character alone then classify the character's image \cite{casey1996survey}, to segmentation free approaches that tried to recognize all the characters in a word, without requiring / performing any explicit segmentation \cite{plotz2009markov}. Today, state-of-the-art text recognition systems work on a whole input line image without requiring any prior explicit character / word segmentation \cite{yousef2018accurate,michael2019evaluating}. This removes the requirement for providing character localization annotations as part of ground-truth transcription. Also the recognition accuracy relies only on automatic line segmentation, a much easier process than automatic character segmentation.

However, line segmentation is still an error-prone process and can cause great deterioration in the performance of today's text recognition systems. This is especially true for documents with hard to segment text-lines such as handwritten documents \cite{gatos2014ground,sanchez2017icdar2017}, with warped lines, uneven interline spacing, touching lines, and torn pages.

\begin{table}[t]

\begin{center}
\resizebox{\columnwidth}{!}{%
\begin{tabular}{C{3.0cm}|c|c|c|c|c} 
\hline
Requirement & \cite{bluche2017scan}& \cite{bluche2016joint}& \cite{tensmeyer2019training}& \cite{chung2019,wigington2018start,moysset2018learning} & Ours \\
\hline
\hline
Full-page image        &\cmark & \cmark & \cmark & \cmark  & \cmark \\
Full-page text GT      &\cmark & \cmark & \cmark &\cmark  & \cmark \\
Seg. line images     &\xmark & \xmark & \xmark & \cmark  & \xmark \\ 
Seg. transcription     &\xmark & \xmark & \xmark & \cmark  & \xmark \\
Pre-train on seg. data &\cmark & \cmark & \cmark & \xmark  & \xmark \\
Special curriculum     &\cmark & \cmark & \xmark & \xmark  & \xmark \\
\# Iterations / image     & 500   & 10 & 10     & 10      & 1 \\
\hline
\end{tabular}
}
\end{center}
\caption{Comparison of what data is required to train a full page recognizer between various prior works and our proposed method. We can see that our method is the only that truly works at page level without requiring any segmented data at any stage. \emph{\# Iterations / image} is the average number of iterations required to transcribe a full paragraph image from the IAM dataset; we can note that while all other methods require multiple iterations per image (to recognize each segmented character or line), our method performs only one pass over the input full paragraph image.}
\label{table:reqcmp}
\end{table}

The main previous works that tried to address the problem of weakly supervised multi-line recognition were \cite{bluche2016joint,bluche2017scan,tensmeyer2019training}. Besides these methods, other methods that work on full page recognition require the localization ground-truth of text lines during training. A detailed comparison between the training data required by our proposed method vs. other methods in literature is presented in Table \ref{table:reqcmp}.

In this work, we present a simple and novel neural network sub-module, termed OrigamiNet, that can be added to any existing convolutional neural network (CNN) text-line recognizer to convert it to a full page recognizer. It can transcribe full text pages in a weakly supervised manner without being given any localization ground-truth (either visual in the images or textual in the transcriptions) during training, and without performing any explicit segmentation. In contrast to previous work, this is done very efficiently using feed-forward connections only (no recurrent connections), essentially, in a single network forward pass. 

Our main intuition in this work is, instead of the traditional two-step framework that first segments then recognizes extracted segments, to propose a novel integrated approach for learning to simultaneously implicitly segment and recognize. This works by learning a representation transformation that \textit{transforms the input into a representation where both segmentation and recognition is trivial}. 

We implicitly unfold an input multi-line image into a single line image (i.e. from a 2D arrangement of characters to 1D), where all lines in the original image are stitched together into one long line, so no text-line segmentation is actually needed. Both segmentation and recognition are done in the same single step (single network forward pass) instead of being carried out iteratively (on each line), and thus all computations are shared between recognition and implicit segmentation, and the whole process is a lot faster.

The main ingredients to achieving this are: Using the idea of a spatial bottleneck followed by up-sampling, used widely in pixel-wise prediction tasks (e.g. \cite{long2015fully,ronneberger2015u}); and using the CTC loss function \cite{graves2006connectionist} which strongly induces / encourages a linear 1D target. We construct a simple neural network sub-module that applies these novel ideas, and demonstrate both its effectiveness and generality by attaching it to a number of state-of-the-art text recognition neural network architectures. We show that it can successfully convert them from single line into multi-line text recognizers with exactly the same training procedure (i.e. without resorting to complex and fragile training recipes, like a special training curriculum or special pre-training strategies).

On the challenging ICDAR 2017 HTR \cite{sanchez2017icdar2017} full page benchmark we achieve state-of-the-art Character Error Rate (CER) without any localization data. On full paragraphs of the IAM \cite{marti2002iam} dataset, we were able to achieve state-of-the-art CER surpassing models that work on carefully pre-segmented text-lines, without using any localization information during training or testing.

To summarize, we address the problem of weakly supervised full-page text recognition. In particular, we make the following contributions:
\begin{itemize}[noitemsep,topsep=0pt,parsep=0pt,partopsep=0pt]
  \item We conceptually propose a new approach for weakly-supervised simultaneous object segmentation and recognition, and apply it to text.
  \item We propose a simple and generic neural network sub-module that can be added to any CNN-based text line recognizer to convert it into a multi-line recognizer that utilizes the same simple training procedure.
  \item We carry an extensive set of experiments on a number of state-of-the-art text recognizers that demonstrate our claims. The resultant architectures demonstrate state-of-the art performance on ICDAR2017 HTR and the full paragraph IAM datasets.
\end{itemize}

\section{Related Work}
There is not much prior work in the literature regarding full page recognition. Segmentation-free multi-line recognition has been mainly considered in \cite{bluche2016joint,bluche2017scan}. The idea of both is using selective attention to focus only on a specific part of the input image, either characters in \cite{bluche2017scan} or lines in \cite{bluche2016joint}. These works have two major drawbacks. First, both are difficult to train, and need to pre-train their encoder sub-network on single-line images before training on multi-line versions, which defeats the objective of the task. Second, though \cite{bluche2016joint} is much faster than \cite{bluche2017scan}, both are very slow compared to current methods that work on segmented text lines. 

Besides these two segmentation-free methods, other methods that work on full page recognition either require the localization ground-truth of text lines for all \cite{carbonell2019,chung2019,moysset2018learning} or part \cite{wigington2018start} of the training data to train either a separate network or a sub-module (of a large, multi-task network) for text-line localization. Also, all these methods require line breaks to be annotated on all the provided textual ground-truth transcriptions (i.e. text lines must be segmented both visually in the image and textually in the transcription). \cite{tensmeyer2019training} presented the idea of adapting  \cite{wigington2018start} in a weakly supervised manner without requiring line breaks in the transcription by setting the alignment between the predicted line transcriptions and the ground truth as a combinatorial optimization problem, and greedily solving it. However \cite{tensmeyer2019training} still requires the same pre-training as \cite{wigington2018start} and performs worse.

\section{Methodology}
\begin{figure*}
\begin{subfigure}{\textwidth}
\centering
\includegraphics[width=0.8\textwidth]{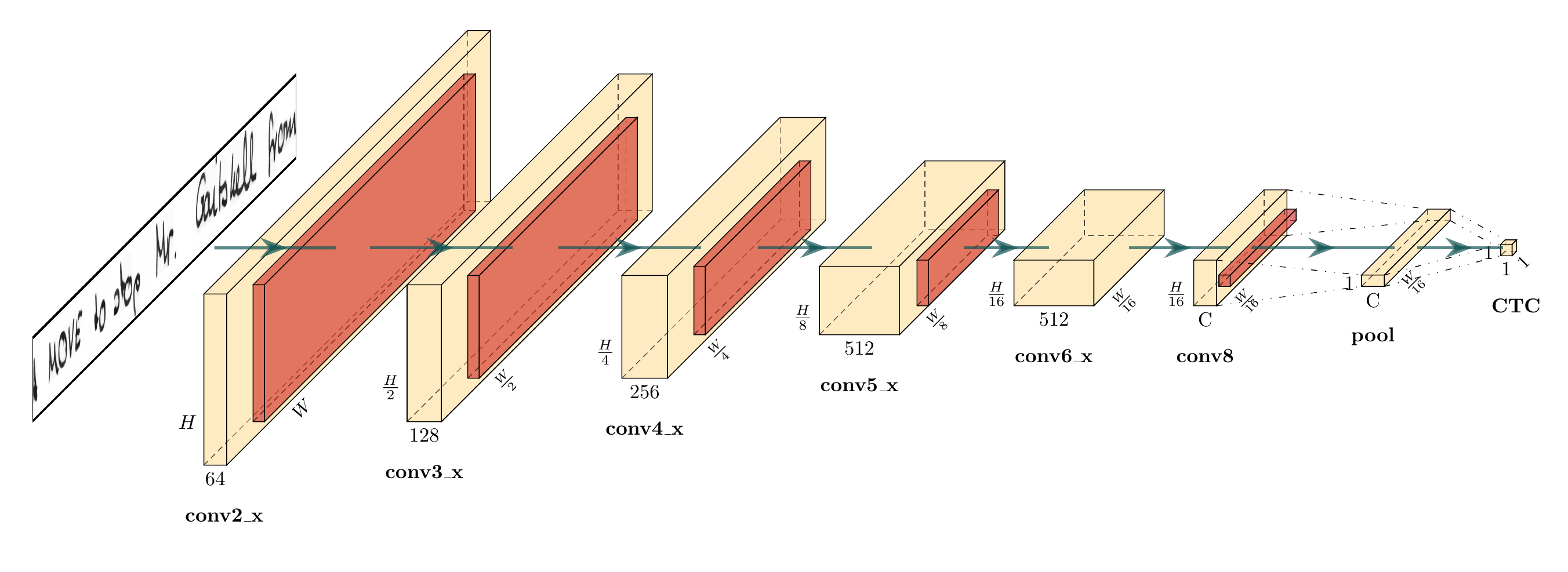}
\caption{A generic four stage fully convolutional single line recognizer, input is a singe line image, training is done using the CTC loss function. Backbone CNN can be any of the ones presented in Table \ref{table:arch}. Input gets progressively down-sampled, then converted into 1D by average pooling along the vertical dimension right before the loss calculation. (Figures created via PlotNeuralNet \cite{haris_iqbal_2018_2526396})}
\label{figa:sl}
\end{subfigure}

\begin{subfigure}{\textwidth}
\centering
\includegraphics[width=0.98\textwidth]{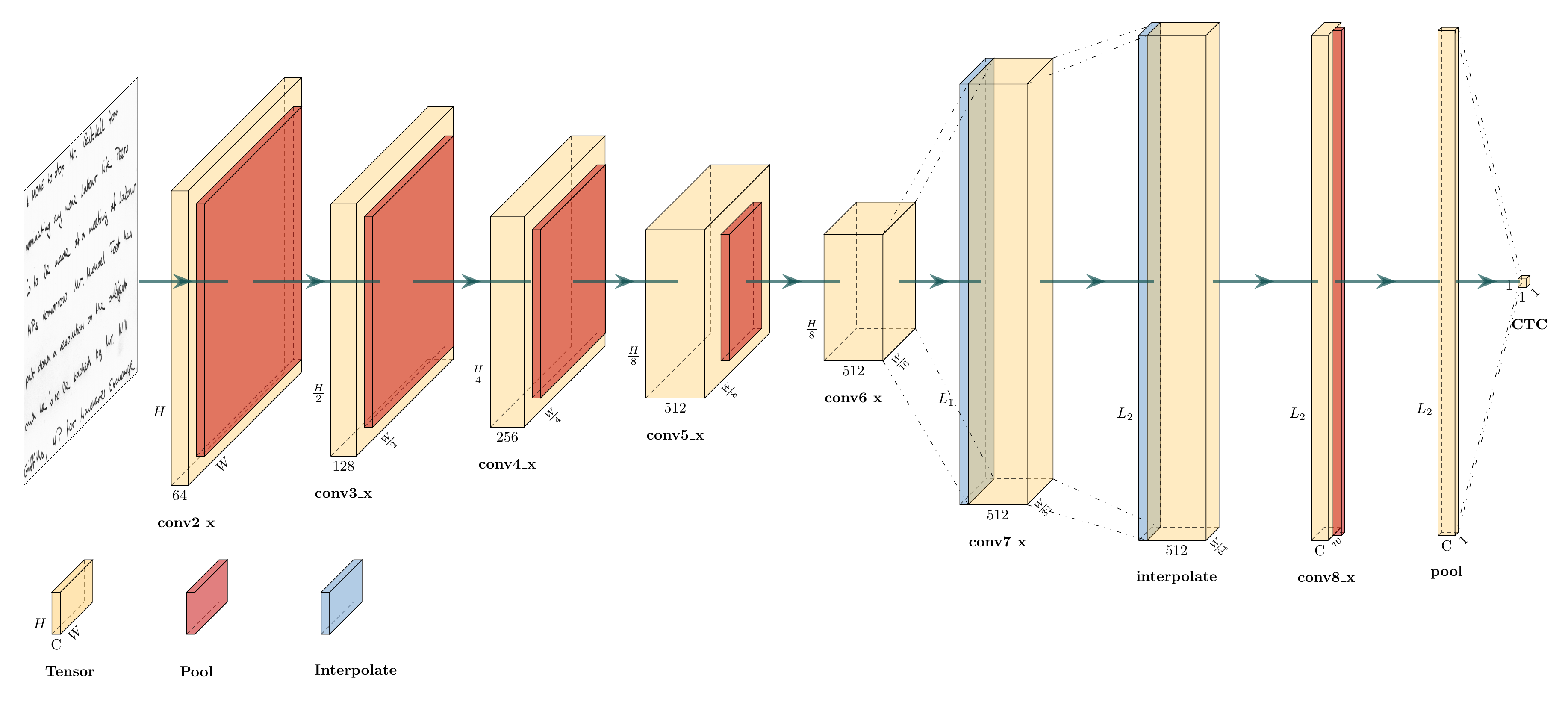}
\caption{Here we convert the fully convolutional single-line recognizer into an OrigamiNet multi-line recognizer; comparing the two figures shows that the main change introduced is up-scaling \emph{vertically} in two stages, and at the same time, down-scaling horizontally. We obtain a feature-map that is \emph{tall} and narrow (the shape of one very long vertical line, length $L_2$). After that we proceed exactly as above, average pooling over the short dimension, $w$ (of the new line \emph{not} the original image) then using the CTC loss function to drive the training process.}
\label{figb:ml}
\end{subfigure}
\caption{Converting a fully-convolutional single line recognizer into a multi-line recognizer using our OrigamiNet module. }
\label{fig:arch}
\end{figure*}

Figure \ref{fig:arch} presents the core idea of our proposed OrigamiNet module, and how it can be attached to any fully convolutional text recognizer. Both before and after versions are shown for easy comparison.

The Connectionist Temporal Classification (CTC) loss function allows the training of neural text recognizers on unsegmented inputs by considering all possible alignments between two 1D sequences. The sequence of predictions produced by the network is denoted $P$, and the sequence of labels associated with the input image $L$, where $|L|<|P|$. The strict requirement of having $P$ as a 1D sequence, introduces a problem, given that the original input signal (the image $I$) is a 2D signal. This problem has typically been dealt with by unfolding the 2D signal into 1D, using a simple reduction operation (e.g. summation) along one of the dimensions (usually the vertical one), giving:
\begin{equation}
    P_i = \sum_{j=1}^H{F(I_{i,j})}
    \label{eq1}
\end{equation}
Where $F$ is a learned 2D representation transformation. This is the paradigm shown in Fig.~\ref{figa:sl}. As noted in \cite{bluche2016joint,bluche2017scan} this simple, \emph{blind} collapse from 2D to 1D gives equal importance / contribution (and therefore gradients) to all the rows of the 2D input feature-map $F(I)$, and thus prevents the recognition of any 2D arrangement of characters in the input image. If two characters cover the same columns, only one can be possibly recognized after the collapse operation.

To tackle this problem, i.e. satisfy the 1D input requirement of CTC without sacrificing the ability of recognizing 2D arrangements of characters, we propose the idea of learning the proper 2D$\rightarrow$1D unfolding through a CNN, motivated by the success of CNNs in pixel-wise prediction and image-to-image translation tasks.

The main idea of our work (presented in Fig.~\ref{figb:ml}) is augmenting the traditional paradigm with a series of up-scaling operations that transforms the input feature-map into the shape of a single line, that is long enough to hold all the lines (2D character arrangements) from the input image. Up-scaling operations are followed by convolutional computational blocks as our learned resize operations (as done by many researchers, e.g. \cite{dong2015image}).  The changed direction of up-scaling encourages each line of the input image to be mapped into a distinct part of the output vertical dimension.

After such changes, we proceed with the traditional paradigm as-is, perform the simple sum reduction (Eq.~\ref{eq1}) along the vertical dimension $w$ of the resulting line (which is perpendicular to the original input multi-line image's vertical dimension). The  model is trained with CTC.

Moreover, we argue that the main bottleneck preventing all previous works from learning proper 2D$\rightarrow$1D mappings directly as we do, is \emph{spatial} constraints (i.e. not overall capacity or architectural constraints). Providing enough spatial capacity to the model allows it to easily learn such transformations (even for simple limited capacity models, as we will show in the experiments section). 
Given the spatial capacity and the strong linear prior induced by CTC, the model is able to learn strong 2D$\rightarrow$1D unfolding function with the same simple training procedure used for training single line recognizers, and without any special pre-training or curriculum applied to any sub-module of the network (both of which are used exclusively in the literature).

One natural question here is how to choose the final line length $L_2$ (see definition in Fig.~\ref{figb:ml})? To gather space for the whole paragraph / page, $L_2$ must be at least as long as the largest number of characters in any transcription in the training set. Longer still is better, given that (i) CTC needs to insert blanks to separate repeated labels; (ii) characters vary greatly in spatial extent, and mapping each to multiple target frames in the final vector is an easier task than transforming to exactly one frame. 

\newcommand{\ve}[1]{\mathbf{#1}} 
\newcommand{\ma}[1]{\mathrm{#1}} 

\newcommand{\tabincell}[2]{\begin{tabular}{@{}#1@{}}#2\end{tabular}}

\renewcommand\arraystretch{1.2}

\def\httilde{\mbox{\tt\raisebox{-.5ex}{\symbol{126}}}}

\newcommand{\blockaa}[2]{\multirow{3}{*}{\(\left[\begin{array}{c}\text{3$\times$3, #1} \end{array}\right]\)$\times$#2}
}

\newcommand{\blockgb}[2]{\multirow{3}{*}{\(\left[\text{$GateBlock(#1)$}\right]\)$\times$#2}
}

\newcommand{\blocka}[2]{\multirow{3}{*}{\(\left[\begin{array}{c}\text{3$\times$3, #1}\\[-.1em] \text{3$\times$3, #1} \end{array}\right]\)$\times$#2}
}
\newcommand{\blockb}[3]{\multirow{3}{*}{\(\left[\begin{array}{c}\text{1$\times$1, #2}\\[-.1em] \text{3$\times$3, #2}\\[-.1em] \text{1$\times$1, #1}\end{array}\right]\)$\times$#3}
}
\renewcommand\arraystretch{1.1}
\setlength{\tabcolsep}{3pt}
\begin{table*}[t]
\begin{center}
\resizebox{0.99\linewidth}{!}{
\begin{tabular}{c|c|c|c|c|c|c|c|c}

\hline
part & layer name & output size & ResNet-26 & ResNet-66 & ResNet-74 & VGG & GTR-8 & GTR-12 \\
\hline
\multirow{20}{*}{\rotatebox{90}{\emph{\Huge Encoder}}} & Input & $H \times W$ & \multicolumn{6}{c}{}\\
\cline{2-9}
& ln1 & $H \times W$  & \multicolumn{6}{c}{static layer normalization} \\
\cline{2-9}
& conv1 & $H \times W$ & \multicolumn{3}{c|}{7$\times$7, 64} & & \multicolumn{2}{c}{ 13$\times$13, 16}\\
\cline{2-9}
& \multirow{4}{*}{conv2\_x} & \multirow{4}{*}{$\frac{H}{2} \times \frac{W}{2}$}  & \blocka{64}{1}  & \blocka{64}{1} & \blocka{64}{1} & \blockaa{64}{1} & \blockgb{512}{1} & \blockgb{512}{1}\\
 & &  &  &  &  &  & &\\
 & &  &  &  &  &  & &\\\cline{4-9}
 & &  &  \multicolumn{6}{c}{2$\times$2 max pool, stride 2} \\
\cline{2-9}
& \multirow{4}{*}{conv3\_x} &  \multirow{4}{*}{$\frac{H}{4} \times \frac{W}{4}$}  & \blocka{128}{2}  & \blocka{128}{2}  & \blocka{128}{6}  & \blockaa{128}{1}  & \blockgb{512}{1}&\blockgb{512}{1}\\
 & &  &  &  &  &  & &\\
 & &  &  &  &  &  & &\\\cline{4-9}
 & &  &  \multicolumn{6}{c}{2$\times$2 max pool, stride 2} \\
\cline{2-9}
& \multirow{4}{*}{conv4\_x} & \multirow{4}{*}{$\frac{H}{8} \times \frac{W}{8}$}  & \blocka{256}{5}  & \blocka{256}{25}  & \blocka{256}{25}  & \blocka{256}{1} & \blockgb{512}{1}& \blockgb{512}{2}\\
 & &  &  &  &  & &\\
 & &  &  &  &  & &\\\cline{4-9}
 & &  &  \multicolumn{6}{c}{2$\times$2 max pool, stride 2} \\
\cline{2-9}
& \multirow{4}{*}{conv5\_x} & \multirow{4}{*}{$\frac{H}{8} \times \frac{W}{16}$}  & \blocka{512}{3}  & \blocka{512}{3}  & \blocka{512}{3}  & \blocka{512}{1} & \blockgb{1024}{1} & \blockgb{1024}{3}\\
 & &  &  &  &  &  & &\\
 & &  &  &  &  &  & &\\\cline{4-9}
 & &  &  \multicolumn{6}{c}{2$\times$2 max pool, stride 1$\times$2} \\
\cline{2-9}
& \multirow{3}{*}{conv6\_x} & \multirow{3}{*}{$\frac{H}{8} \times \frac{W}{16}$}  & \blocka{512}{1}  & \blocka{512}{1}  & \blocka{512}{1}  & \blocka{512}{1} & \blockgb{1024}{3} & \blockgb{1024}{4}\\
 & &  &  &  &  &  & &\\
 & &  &  &  &  &  & &\\
\hline
\specialrule{.3em}{.2em}{.2em}
\hline
\multirow{10}{*}{\rotatebox{90}{\emph{\Huge Decoder}}} & \multirow{4}{*}{conv7\_x} & \multirow{4}{*}{$L_1 \times \frac{W}{32}$} & \multicolumn{5}{c}{interpolate bilinearly to $L_1 \times \frac{W}{32}$ } \\\cline{4-9}
& & & \blocka{512}{3}  & \blocka{512}{3}  & \blocka{512}{3}  & \blocka{512}{1} & \blockgb{512}{1} & \blockgb{512}{1}\\
&  &  &  &  &  &  & &\\
&  &  &  &  &  &  & &\\
\cline{2-9}
& & $L_2 \times \frac{W}{64}$  & \multicolumn{6}{c}{interpolate bilinearly to $L_2 \times \frac{W}{64}$} \\
\cline{2-9}
& conv8 & $L_2 \times w$  & \multicolumn{6}{c}{1$\times$1, $C$} \\
\cline{2-9}
& & $L_2$  & \multicolumn{6}{c}{average pool over short dimension $w$} \\
\cline{2-9}
& ln2 & $L_2$  & \multicolumn{6}{c}{static layer normalization} \\
\cline{2-9}
& & 1  & \multicolumn{6}{c}{CTC} \\
\hline
\multicolumn{3}{c|}{\# Parameters $\times 10^6$} & 38.2  & 61.9  & 63.05  & 10.6  & 9.9 & 16.4 \\
\hline
\end{tabular}
}
\end{center}
\vspace{-.5em}
\caption{Architectural details of our evaluated CNN backbones (\emph{Encoder} part), and how our module (\emph{Decoder} part) is attached to them. The table tries to abstract the architectures to their most common details. Although there is subtle difference in the components of the basic building block (in brackets [])  of every architecture, the overall organization of the network, and how our module fits, is the same.}
\label{table:arch}
\vspace{-.5em}
\end{table*}

\section{Experiments}
We carry out an extensive set of experiments to answer the following set of questions:
\begin{itemize}[noitemsep,topsep=0pt,parsep=0pt,partopsep=0pt]
\item Does the module actually work as expected?
\item Is it tied to a specific CNN architecture?
\item Is it tied to a specific model capacity?
\item How does final spatial size affect model performance?
\end{itemize}

\subsection{Implementation Details}
All experiments use an initial learning rate of 0.01, exponentially decayed to 0.001 over  $9\times 10^4$ batches. We implement in PyTorch \cite{NEURIPS2019_9015}, with the Adam \cite{Kingma:2014usa} optimizer.

\subsection{Datasets}
IAM \cite{marti2002iam} (modern English) is a famous offline handwriting benchmark dataset. It is composed of 1539 scanned text pages  handwritten by 657 different writers, corresponding to English texts extracted from the LOB corpus \cite{Johansson1980TheLC}. IAM has 747 documents (6,482 lines) in the training set, 116 documents (976 lines) in the validation set and 336 documents (2,915 lines) in the test set.

The ICDAR2017 full page HTR competition \cite{sanchez2017icdar2017} consists of two training sets. The first contains 50 fully annotated images with line-level localization and transcription ground-truth. The second set contains 10,000 images with only transcriptions (with annotated line breaks). Most of the dataset was taken from the Alfred Escher Letter Collection (AEC) which is written in German but it also has pages in French and Italian. In all our experiments on this dataset, \emph{we don't make any use of} either the 50-page training set or the annotated line-breaks on the 10,000-page training set

\subsection{CNN Backbones}
To emphasize the generality of our proposed module, we evaluate it on a number of popular CNN architectures that achieved strong performance in the text recognition literature. Inspired by the benchmark work \cite{baek2019wrong}, we evaluate VGG and ResNet-26 (the specific variants explored in \cite{baek2019wrong}), as well as deeper and much more expressive variants (ResNet-66 and ResNet-74). We also evaluate a newly proposed gated, fully convolutional architecture for text recognition \cite{yousef2018accurate}, named Gated Text Recognizer (GTR). The detailed structure of the CNN backbones we evaluate our proposed model on is presented in Table \ref{table:arch}. More details on the basic building blocks of these architectures can be found in their respective papers, VGG \cite{simonyan2014very}, ResNet \cite{he2016deep}, and GTR \cite{yousef2018accurate}.

\subsection{Final Length, $L_2$}
For IAM, the final length should be at least 625, since the longest paragraph in the training set contains 624 characters. We have two questions here: what value can balance running time and recognition accuracy? And how does the relation between $L_1$ and $L_2$ affect the final CER?

Table \ref{table:fwc} presents some experiments on this. First, we can see that generally, even a very simple model like VGG can successfully learn to recognise multiple lines (at a relatively bad CER = 30\%) at various configurations, yet, the deeper ResNet-26 achieves a much better performance on the task reaching 7.2\%. Second, it is evident that wider generally gives better performance (but at diminishing returns), which is evident for VGG more than ResNet-26. We see that for reasonable values (>800) the network is fairly robust to the choice of $L_2$. We can also note that both $L_1$ and $L_2$ should be relatively close to each other.

\subsection{Final Width}
Does the final shape need to have the largest possible aspect ratio? How would the final width, $w$ (shorter output dimension) affect the learning system?
Table \ref{table:fhc} presents experiments using VGG and ResNet-26 on this regard. It is clear that a large value like 62 deteriorates training significantly for ResNet-26, but small and medium values (<31) are comparable in performance. On the other hand, a model with limited receptive field and complexity like VGG can generally make a lot of use from the added width.

\subsection{End-to-end Layer Normalization}

The idea of using parameter-less layer normalization as the first and last layer of a model was proposed in \cite{yousef2018accurate}, and shown to increase performance and facilitate optimization. The same idea was very effective for our module, as initially some deep models that converged for single line recognition completely diverged here. This is most probably due to the large number of time-steps CTC works on for our case.

As can be seen in Table \ref{tbl:fnl}, end-to-end layer normalization can bring significant increases in accuracy for models that already worked well; more importantly, it makes it possible to train very deep models that were constantly diverging before, leading to state-of-art performance on the task.

\subsection{Hard-to-segment text-lines}
\begin{figure*}
\centering
\begin{subfigure}{0.23\textwidth}
\includegraphics[width=\textwidth]{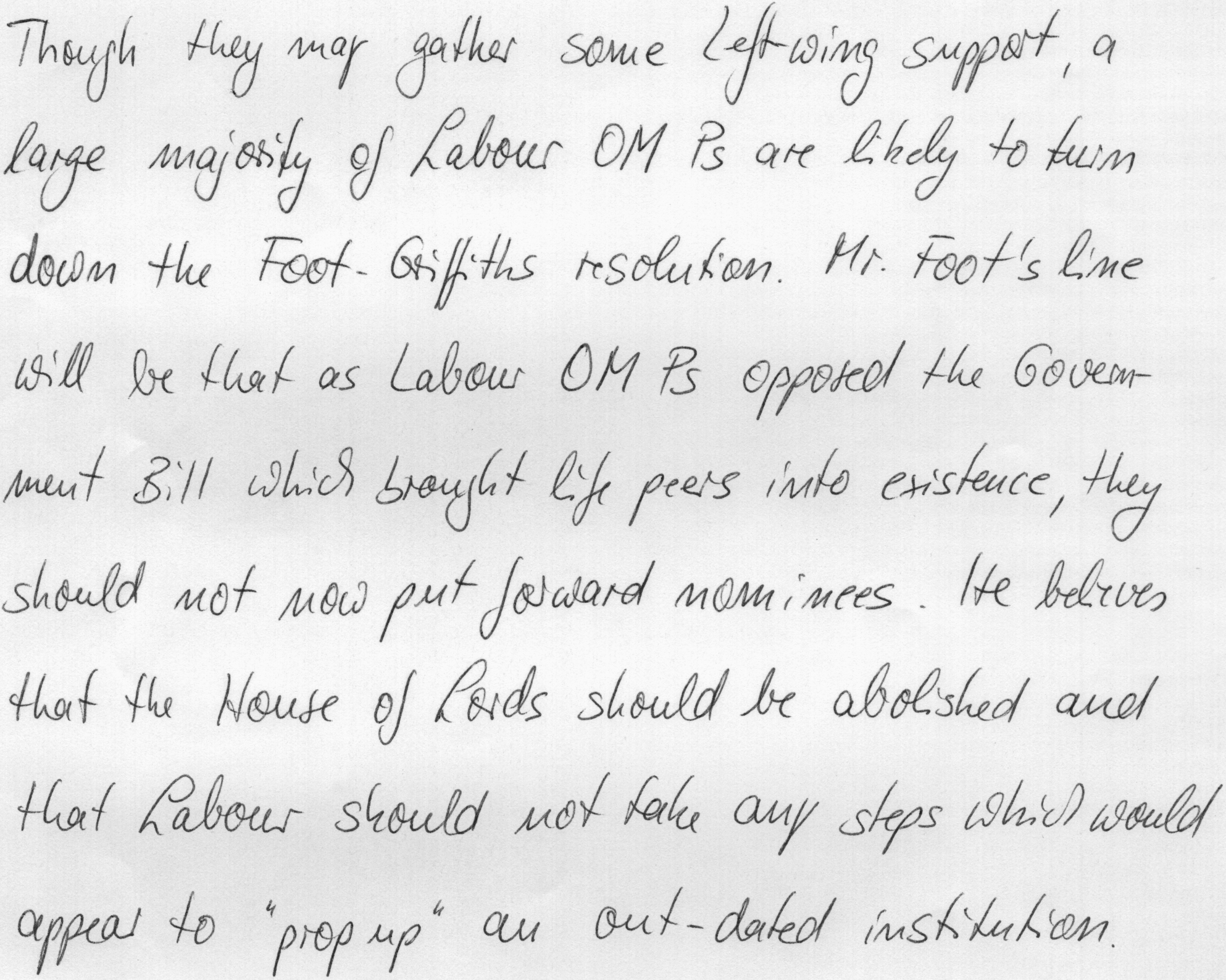}
\end{subfigure}
\begin{subfigure}{0.23\textwidth}
\includegraphics[width=\textwidth]{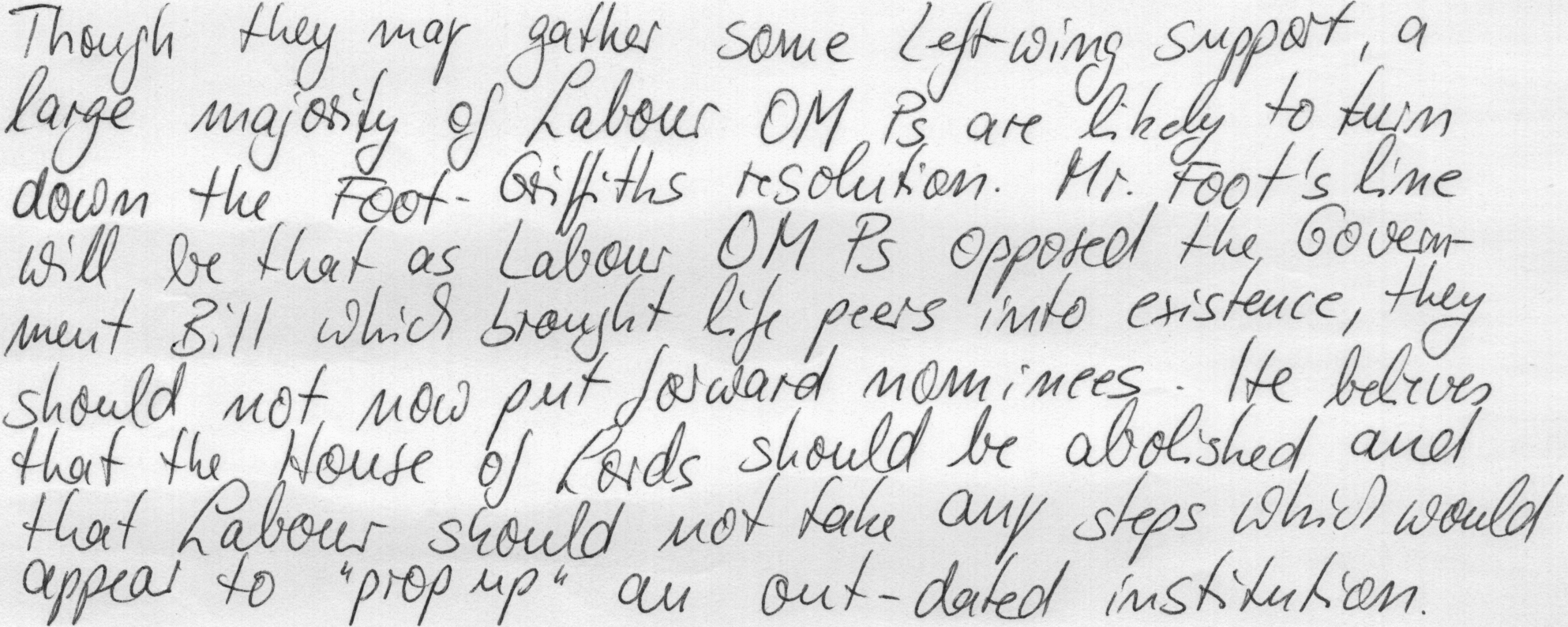}
\end{subfigure}
\begin{subfigure}{0.23\textwidth}
\includegraphics[width=\textwidth]{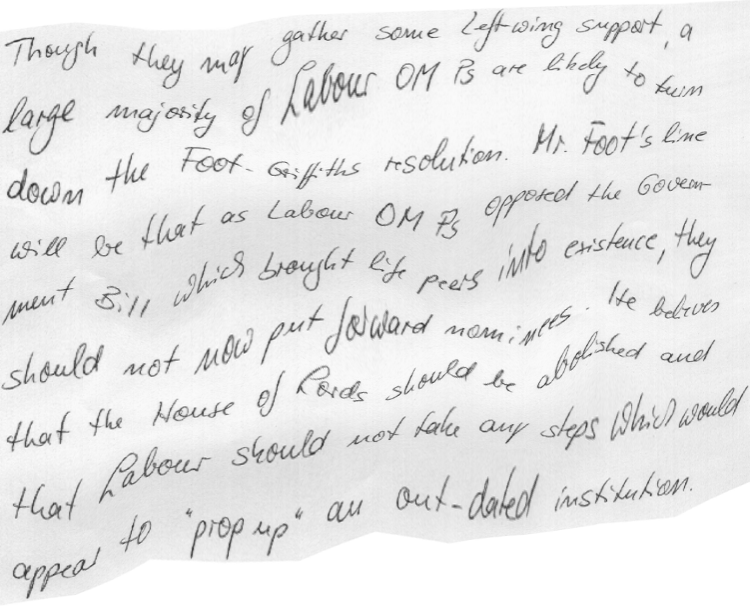}
\end{subfigure}

\begin{subfigure}{0.23\textwidth}
\includegraphics[width=\textwidth]{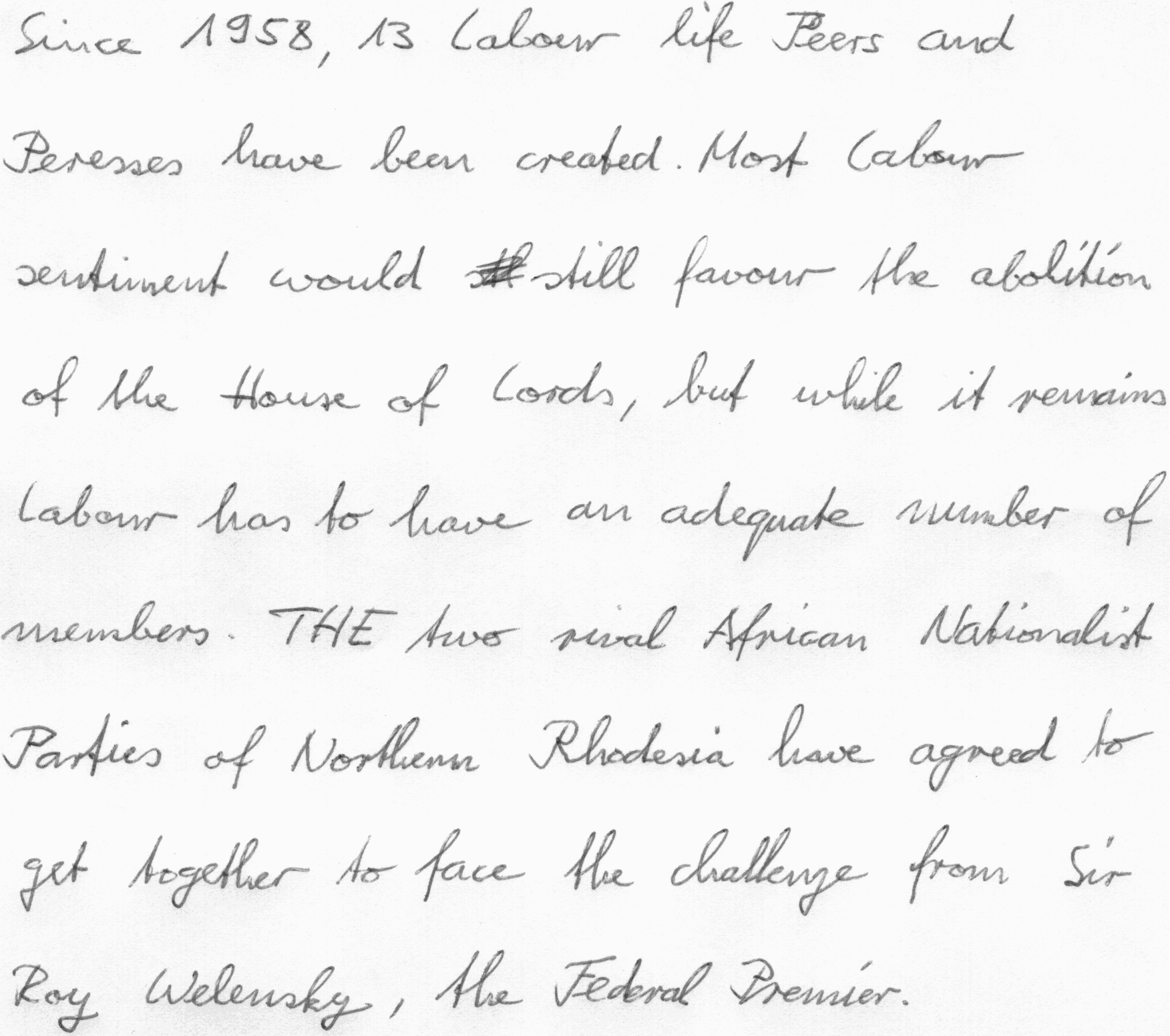}
\caption{{Original Image.}}
\end{subfigure}
\begin{subfigure}{0.23\textwidth}
\includegraphics[width=\textwidth]{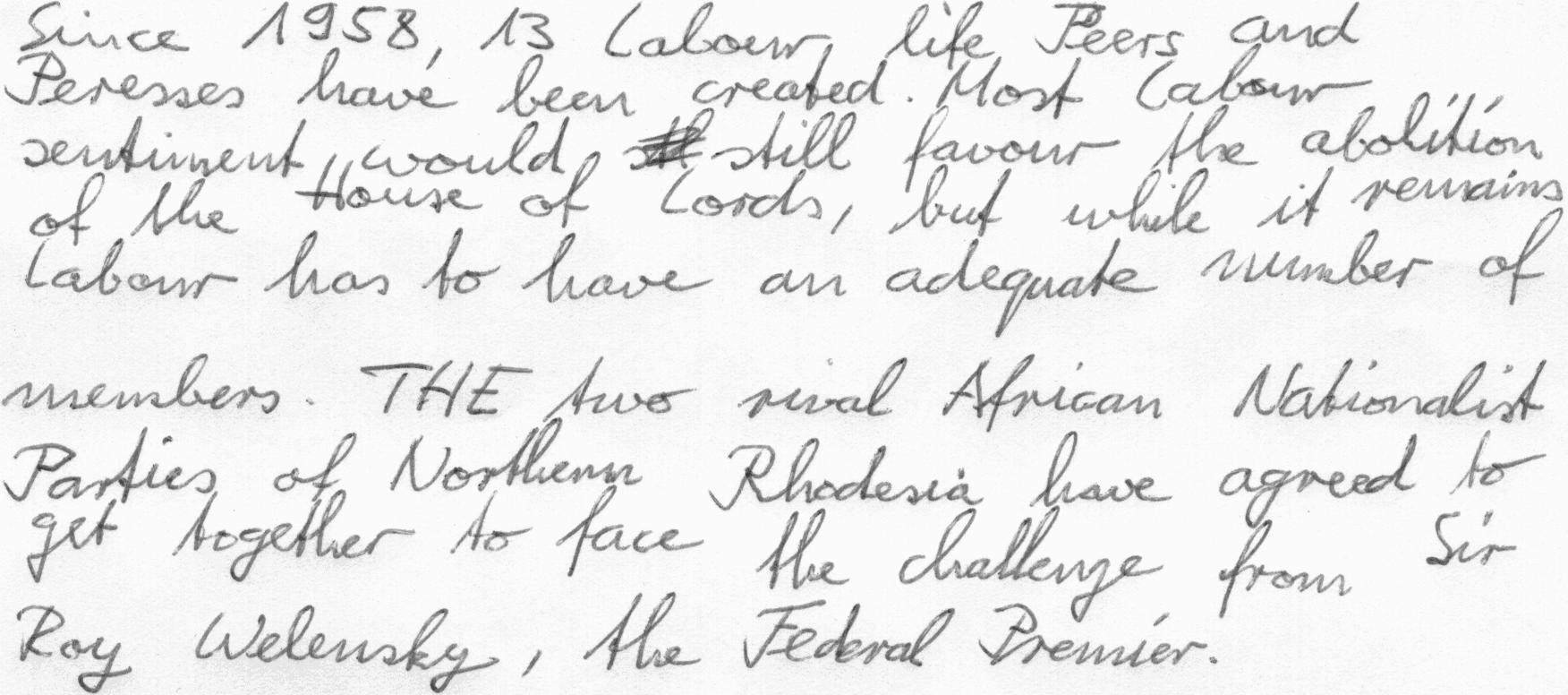}
\caption{{Compact lines.}}
\end{subfigure}
\begin{subfigure}{0.23\textwidth}
\includegraphics[width=\textwidth]{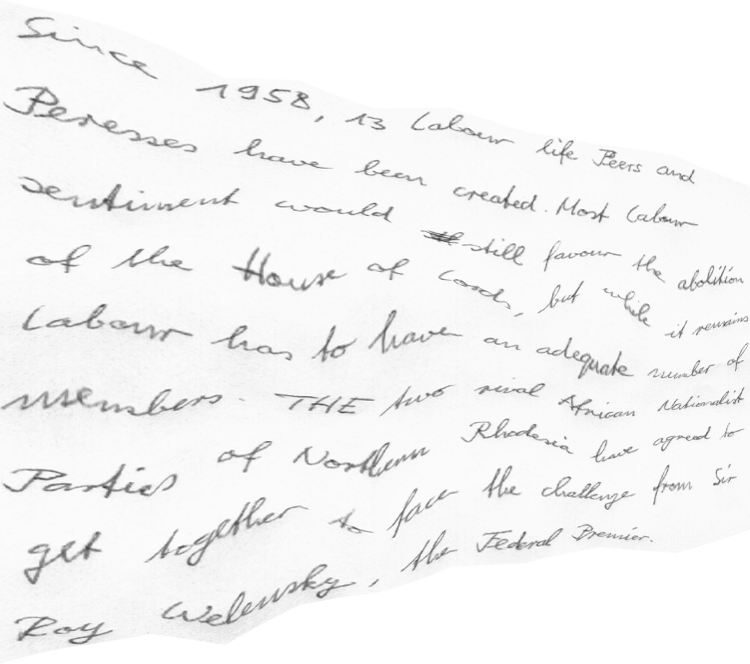}
\caption{{Rotated and warped.}}
\end{subfigure}

\caption{Synthetic distortions applied to the IAM dataset to study the how our model handles hard to segment text-lines. (a) original paragraph image. (b) touching text-lines. (c) rotated and wavy text-lines \label{fig:hard}}
\end{figure*}

Due to the way IAM was collected \cite{marti2002iam}, its lines are generally easy to segment. 
To study how our model would handle harder cases, we carried out two separate experiments, artificially modifying IAM to produce new variants with hard-to-segment lines. Firstly, interline spacing is massively reduced via seam carving \cite{avidan2007seam}, resizing to 50\% height, creating heavily touching text lines, Fig.~\ref{fig:hard}(b). GTR-12 achieved 6.5\% CER on this dataset.
Secondly, each paragraph has random projective (rotating/resizing lines), and random elastic transforms (like \cite{wigington2017data} but at the page level) applied, creating wave-like non-straight lines, Fig.~\ref{fig:hard}(c). GTR-12 achieved 6.2\% CER on this dataset.

\subsection{Comparison to state-of-the-art}
For all the previous experiments, IAM paragraph images were scaled down to 500 $\times$ 500 pixels before training, and although we were already achieving state-of-the-art results, we wanted to explore whether we can break even with single line recognizers.
As shown in Table~\ref{tbl:sota}, by increasing image / model sizes, we were for the first time able to exceed the performance of state-of-the-art single line recognizers using a segmentation free full page recognizer that trains without any visual or textual localization ground-truth. Note that we don't include in the comparison methods that use additional data, either in the form of training images as in \cite{xiao2019deep,dutta2018improving} or language modeling as in \cite{voigtlaender2016handwriting}.

For the ICDAR2017 HTR dataset we follow \cite{tensmeyer2019training} and report CER on the validation set proposed in \cite{wigington2018start} (the last 1000 pages of the 10,000 image training set), as the evaluation server doesn't provide CER or other character based metrics. Results are in Table~\ref{table:icdar}. Note that both \cite{wigington2018start,tensmeyer2019training} report results using CER normalized by GT length (\emph{nCER} in the table). We used author released pre-trained models from \cite{wigington2018start} to compute their results without a language model. It is very evident our method can get far superior performance using weaker training signals. 

\subsection{Model Interpretability}
Here we consider an important question: what does the model actually learn? We can see that the model works well in practice and we have a hypothesis of what it \emph{might} be doing, but it would very interesting if we can have a peek at how our model is able to make its predictions.

To gain an understanding of what parts of the input biases the model towards a specific prediction, we utilize the framework of Path-Integrated Gradients \cite{sundararajan2017axiomatic} ensembled using SmoothGrad \cite{smilkov2017smoothgrad}. Note that unlike typical classification tasks, we predict $L_2$ labels per image. Of those we discard blanks and repeated consecutive labels (in CTC, representing continuation of the same state; we found their attribution maps to be global and uninformative for these purposes).

For integrated gradients (IG), we change the baseline to use an empty white image to designate no-signal, rather than an empty black one (which would be an all-signal image in our case) - as our data is black text over a white background. Using white baselines produced much sharper attribution maps than black ones, showing how sensitive IG is to the choice of the baseline (studied more in \cite{sturmfels2020visualizing}). We used 50 steps to approximate the integral in our tests.

Standard SmoothGrad produces attribution maps that are very noisy (see \cite{sturm2016interpretable}), but the SmoothGrad-Squared variant often suppresses most of the signal (a direct consequence of squaring fractions). After analysing the results of both, we suggest the root cause of SmoothGrad problems is averaging positive and negative signals together. The squaring in SmoothGrad-Squared solves this problem, but at the cost of suppressing some important parts of the signal. So we propose \emph{SmoothGrad-Abs}, which simply averages the absolute value of the attribution maps. SmoothGrad-Abs strikes a good balance between SmoothGrad and SmoothGrad-Squared. For our experiments, we used 5 noisy images.

\begin{figure*}
\begin{subfigure}{0.23\textwidth}
\includegraphics[width=\textwidth]{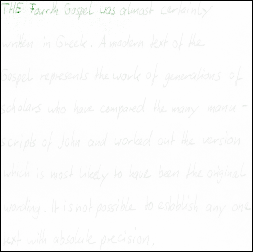}
\caption*{\scalebox{.60} {THE \colorbox{green}{F}ourth Gospel was almost certainly}}
\end{subfigure}
\begin{subfigure}{0.23\textwidth}
\includegraphics[width=\textwidth]{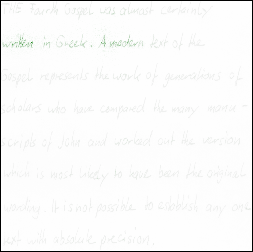}
\caption*{\scalebox{.60} {writt\colorbox{green}{e}n in Greek. A modern text of the}}
\end{subfigure}
\begin{subfigure}{0.23\textwidth}
\includegraphics[width=\textwidth]{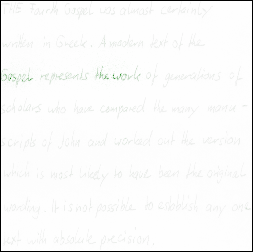}
\caption*{\scalebox{.60} {\colorbox{green}{G}ospel represents the work of generations of}}
\end{subfigure}
\begin{subfigure}{0.23\textwidth}
\includegraphics[width=\textwidth]{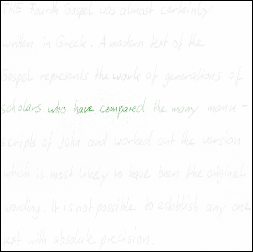}
\caption*{\scalebox{.60} {scholars who hav\colorbox{green}{e} compared the many manu-}}
\end{subfigure}

\begin{subfigure}{0.23\textwidth}
\includegraphics[width=\textwidth]{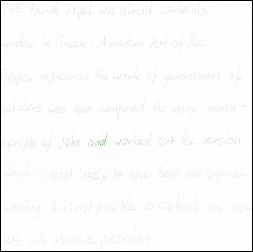}
\caption*{\scalebox{.60} {scripts of John an\colorbox{green}{d} worked out the version}}
\end{subfigure}
\begin{subfigure}{0.23\textwidth}
\includegraphics[width=\textwidth]{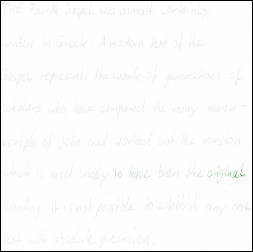}
\caption*{\scalebox{.60} {which is most likely to have been the origina\colorbox{green}{l}}}
\end{subfigure}
\begin{subfigure}{0.23\textwidth}
\includegraphics[width=\textwidth]{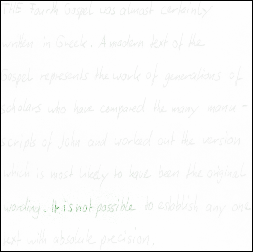}
\caption*{\scalebox{.60} {wording. I\colorbox{green}{t} is not possible to establish any one}}
\end{subfigure}
\begin{subfigure}{0.23\textwidth}
\includegraphics[width=\textwidth]{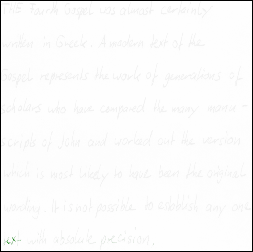}
\caption*{\scalebox{.60} {t\colorbox{green}{e}xt with absolute precision.}}
\end{subfigure}

\caption{Results of the interpretability experiment. For each of these 8 images (from left-right, top-down) we show the attribution heat-map for a single character output (for each line in the image) overlaid over a faint version of the original input image. The randomly chosen character is highlighted in green in the transcription below the image.\label{fig:int1}}
\end{figure*}

Fig.~\ref{fig:int1} shows the attribution maps of a single random character from each line of the input image (computed from the attribution of the corresponding output neuron in the 1D prediction map fed to CTC). We see that the model does indeed implicitly learn good character-level localization from the input 2D image to the output 1D prediction map.

\begin{figure*}
\begin{subfigure}{0.23\textwidth}
\includegraphics[width=\textwidth]{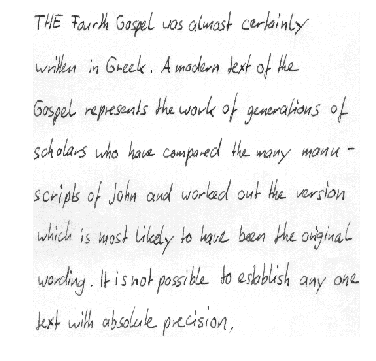}\raisebox{0.5\textwidth}{$\Rightarrow$}
\end{subfigure}
\begin{subfigure}{0.23\textwidth}
\includegraphics[width=\textwidth]{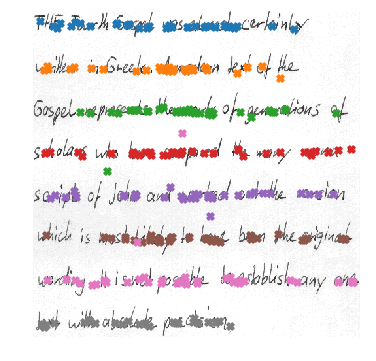}
\end{subfigure}
\begin{subfigure}{0.23\textwidth}
\includegraphics[width=\textwidth]{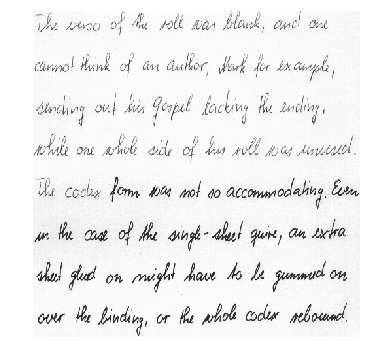}\raisebox{0.5\textwidth}{$\Rightarrow$}
\end{subfigure}
\begin{subfigure}{0.23\textwidth}
\includegraphics[width=\textwidth]{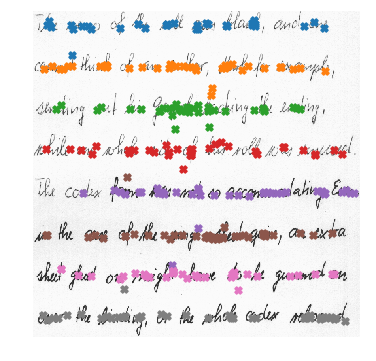}
\end{subfigure}
\caption{\label{fig:int2} The first and third columns represent two input images. The second and fourth columns are the corresponding color coded scatter plot, where, for each character, the position of the center of mass for the attribution map associated with that character is marked. Character markers belonging to the same line are given the same color. We can see that the model learns a very good implicit segmentation of the input image into lines without any localization signal.}
\end{figure*}

Fig.~\ref{fig:int2} provides a holistic view that gathers all the maps into one image. We took the one-character attribution map from the previous step, apply Otsu thresholding to it (to keep only the most important parts) then add a marker at the position of the center of mass of the resulting binary image. The marker is colored according to the transcription text line it belongs to. As can be seen, the result represents a very good implicit line segmentation of the original input.

\begin{table}
\begin{center}
\begin{tabular}{c|c|c|c|c|c} 
\hline
Final length ($L_2$) & 700 & 800 & 950 & 1100 & 1500 \\
\hline
\hline
\multicolumn{6}{c}{First stage length $L_1$ = 450} \\
\hline
VGG & 43.14 & 34.32 & 34.55 & 34.55 & 30.34 \\
\hline
ResNet-26 & 8.121 & 7.675 & 7.602 & 7.238 & 7.449\\
\hline
\multicolumn{6}{c}{First stage length $L_1$ = 225} \\
\hline
VGG &  37.5 & 39.6 & 37.5 & 36.46 & 34.75\\
\hline
\end{tabular}
\end{center}
\caption{The IAM test set CER of VGG and ResNet-26 for various values of $L_1$ and $L_2$.}
\label{table:fwc}
\end{table}

\begin{table}
\begin{center}
\begin{tabular}{c|c|c|c|c|c} 
\hline
Final width & 62 & 31 & 15 & 8 & 3 \\
\hline
\hline
VGG & 25.98 & 17.41  & 37.4 & 34.55 & 24.21\\
\hline
ResNet-26 & 19.9 & 9.128 & 8.64 & 7.238 & 8.34\\
\hline
\end{tabular}
\end{center}
\caption{The IAM test set CER of VGG and ResNet-26 for various final widths. Here $L_1=450$ and $L_2=1100$ }
\label{table:fhc}
\end{table}

\begin{table}
\begin{center}
\begin{tabular}{c|c|c|c|c|c} 
\hline
LN & VGG & ResNet-26 & ResNet-66 & ResNet-74 & GTR-8 \\
\hline
\hline
w/o & 51.37 & 10.03  & 8.925 & 76.9 & 72.4\\
\hline
w & 34.55 & 7.238 & 6.373 & 6.128 & 5.639 \\
\hline
\end{tabular}
\end{center}
\caption{The IAM test set CER for various models, with and without layer-normalization}
\label{tbl:fnl}
\end{table}

\newcommand\sparbox[3][c]{\parbox[#1]{#2}{\centering \strut #3 \strut}}

\begin{table*}
\label{table:fwd}
\begin{center}
\begin{tabular}{c|c|c|c} 
\hline
Method & Input Scale &  Test CER(\%) & Remarks \\
\hline
\hline
\multicolumn{4}{c}{\emph{Single-line methods}} \\
\hline
\cite{puigcerver2017multidimensional} & 128 $\times$ W & 5.8 & CNN+BLSTM+CTC \\
\cite{michael2019evaluating} & 64 $\times$ W & 5.24 & Seq2Seq (CNN+BLSTM encoder) \\
\cite{yousef2018accurate} & 32 $\times$ W & 4.9 & CNN+CTC \\
\hline
\hline
\multicolumn{4}{c}{\emph{Multi-line methods}} \\
\hline
\cite{bluche2017scan}  & 150 dpi & 16.2  & \multirow{3}{*}{\sparbox{0.35\textwidth}{Requires pre-training the encoder (MDLSTM) on segmented text lines}}\\
\cite{bluche2016joint} & 150 dpi & 10.1  & \\
\cite{bluche2016joint} & 300 dpi & 7.9 & \\
\hline
\cite{carbonell2019} & 150 dpi & 15.6 & \multirow{2}{*}{Requires fully segmented training data} \\
\cite{chung2019} & & 8.5 & \\
\hline
\cite{wigington2018start} & & 6.4 & \sparbox{0.35\textwidth}{Requires full line-break annotation and partial visual localization}\\
\hline
ResNet-74 OrigamiNet & 500 $\times$ 500 & 6.1 &\\
GTR-8 OrigamiNet & 500 $\times$ 500 & 5.6 &\\
GTR-8 OrigamiNet & 750 $\times$ 750 & 5.5 &\\
GTR-12 OrigamiNet & 750 $\times$ 750 & \textbf{4.7} &\\
\hline
\hline
\end{tabular}
\end{center}
\caption{Comparison with the state-of-the-art on the IAM paragraph images, best result is highlighted.}
\label{tbl:sota}
\end{table*}

\begin{table}
\begin{center}
\resizebox{\columnwidth}{!}{%
\begin{tabular}{c|c|c|c|c} 
\hline
Method & CER & nCER &  linebreaks & Pre-train \\
\hline
\hline
SFR \cite{tensmeyer2019training} & 8.18 & 8.68 & \cmark & \multirow{2}{*}{\sparbox{0.12\textwidth}{50 fully annotated pgs}}\\
\cline{1-4}
SFR-align \cite{wigington2018start} & - & 11.05 & \xmark \\
\hline
GTR-12 OrigamiNet & \textbf{6.80} & \textbf{5.87} & \xmark & -\\
\hline
\end{tabular}
}
\end{center}
\caption{Comparison on ICDAR2017 HTR, best result is highlighted. \emph{nCER} is CER normalized by GT length. \emph{linebreaks} indicates their presence or removal from the GT.}
\label{table:icdar}
\end{table}

\subsection{Limitations}
We also trained our network on a variant of IAM with horizontally flipped images and line-level flipped groundtruth transcription, where it managed to achieve nearly the same CER. This verifies that the proposed method is robust and can learn the reading order from data.

While the proposed method works well on paragraphs or full pages of text, learning the flow of multiple columns is not addressed directly. However, given that region / paragraph segmentation is trivial compared to text line segmentation we think this is not a serious practical limitation.

\section{Conclusion}
In this paper we tackled the problem of multi-line / full page text recognition without any visual or textual localization ground-truth provided to the model during training. We proposed a simple neural network sub-module, OrigamiNet, that can be added to any existing fully convolutional single-line recognizer and convert it into a multi-line recognizer by providing the model with enough spatial capacity to be able to properly unfold 2D input signals into 1D without losing information.

We conducted an extensive set of experiments on the IAM handwriting dataset to show the applicability and generality of our proposed module. We achieve state-of-the-art CER on the ICDAR2017 HTR and IAM datasets surpassing models that explicitly made use of line segmentation information during training. We then concluded with a set of interpretability experiments to investigate what the model actually learns and demonstrated its implicit ability to localize characters on each line.

{\small
\bibliographystyle{ieee}
\bibliography{egbib}
}

\end{document}